\documentclass[11pt,a4paper]{article}
\usepackage{times,latexsym}
\usepackage{url}
\usepackage[T1]{fontenc}
\usepackage[utf8]{inputenc}

\usepackage[acceptedWithA]{tacl2018v2}

\usepackage{xspace,mfirstuc,tabulary}

\newif\iftaclinstructions
\taclinstructionsfalse 
\iftaclinstructions
\renewcommand{\confidential}{}
\renewcommand{\anonsubtext}{(No author info supplied here, for consistency with
TACL-submission anonymization requirements)}
\newcommand{\instr}
\fi

\iftaclpubformat 

\else

\fi

\usepackage{microtype}
\usepackage{amsmath}
\usepackage{amsfonts}
\usepackage{booktabs}
\usepackage{amsthm}
\usepackage{tipa}
\usepackage[inline, shortlabels]{enumitem}
\setlist{itemjoin ={,\enspace},itemjoin* = { and\enspace}}
\usepackage{xspace}
\usepackage{adjustbox}
\usepackage{textcomp}
\usepackage{array}

\usepackage{todonotes}
\makeatletter
\newcommand*\iftodonotes{\if@todonotes@disabled\expandafter\@secondoftwo\else\expandafter\@firstoftwo\fi}  
\makeatother

\usepackage{cleveref}
\crefname{section}{\S}{\S\S}
\Crefname{section}{\S}{\S\S}
\crefname{table}{Table}{}
\crefname{figure}{Figure}{}
\crefname{algorithm}{Algorithm}{}
\crefname{equation}{Equation}{}
\crefname{appendix}{Appendix}{}
\crefformat{section}{\S#2#1#3}  

\newcommand*{\mybox}[1]{%
  \fcolorbox{white}{white}{\raisebox{0pt}[0.5\baselineskip][0.05\baselineskip]{%
      \hbox to 1.25cm{\hss#1\hss}}}}

\newcommand*{\myboxgrey}[1]{%
  \fcolorbox{gray!25}{gray!25}{\raisebox{0pt}[0.5\baselineskip][0.05\baselineskip]{%
    \hbox to 1.25cm{\hss#1\hss}}}}

\usepackage{tikz}
\usetikzlibrary{bayesnet}

\newcommand{\surface}{\mathbf{s}}
\newcommand{\us}{\mathbf{u}_\mathbf{s}}
\newcommand{\under}{\mathbf{u}}
\newcommand{\uu}{\mathbf{u}}

\newcommand{\vmu}{\boldsymbol \mu}
\newcommand{\veps}{\boldsymbol \epsilon}
\newcommand{\mm}{\mathbf{m}}

\newcommand{\vh}{\mathbf{h}}
\newcommand{\decomp}{\textrm{decomp}}

\newcommand{\defn}[1]{\textbf{#1}}

\newcommand{\word}[1]{\textit{#1}}
\newcommand{\mtag}[1]{\textsc{#1}}

\newcommand{\calP}{{\cal P}}

\newcommand{\pur}{p_{\textit{ur}}}
\newcommand{\psr}{p_{\textit{sr}}}
\newcommand{\pphon}{p_{\textit{phon}}}

\title{Differentiable Generative Phonology}

\newcommand{\jhu}{1}
\newcommand{\mila}{2}
\newcommand{\mcgill}{3}
\newcommand{\ucambridge}{4}
\newcommand{\ethz}{5}

\makeatletter
\newcommand{\printfnsymbol}[1]{%
  \textsuperscript{\@fnsymbol{#1}}%
}
\makeatother

\author{\bf Shijie Wu\thanks{~~Equal contribution}$^{~~\jhu}$~\;~Edoardo M. Ponti$^{*~\mila,\mcgill,\ucambridge}$~\;~Ryan Cotterell$^{~\ucambridge,\ethz}$ \\ 
$^{\jhu}$Johns Hopkins University~\;~$^{\mila}$Mila Montreal~\;~$^{\mcgill}$McGill University \\
$^{\ucambridge}$University of Cambridge~\;~$^\ethz$ETH Z\"urich~\ \\
$^{\jhu}$\texttt {shijie.wu@jhu.edu}~\;~$^{\ucambridge}$\texttt {\{ep490,rdc42\}@cam.ac.uk}
}

\date{}

\begin{document}
\maketitle

\begin{abstract}

The goal of generative phonology, as formulated by \newcite{Chomsky1968}, is to specify a formal system that explains the set of attested phonological strings in a language. Traditionally, a collection of rules (or constraints, in the case of optimality theory) and underlying forms (UF) are posited to work in tandem to generate phonological strings. However, the degree of abstraction of UFs with respect to their concrete realizations is contentious. 
As the main contribution of our work, we implement the phonological generative system as a neural model differentiable end-to-end, rather than as a set of rules or constraints. Contrary to traditional phonology, in our model UFs are continuous vectors in $\mathbb{R}^d$, rather than discrete strings. As a consequence, UFs are discovered automatically rather than posited by linguists, and the model can scale to the size of a realistic vocabulary. Moreover, we compare several modes of the generative process, contemplating: i) the presence or absence of an underlying representation in between morphemes and surface forms (SFs); and ii) the conditional dependence or independence of UFs with respect to SFs. We evaluate the ability of each mode to predict attested phonological strings on 2 datasets covering 5 and 28 languages, respectively. The results corroborate two tenets of generative phonology, viz.\ the 
necessity for UFs and their independence from SFs. In general, our neural model of generative phonology learns both UFs and SFs automatically and on a large-scale. 
The code is available at \url{https://github.com/shijie-wu/neural-transducer}.

\end{abstract}

\section{Introduction}\label{sec:introduction}
Generative phonology is one of the most prominent paradigms in phonological analysis
\cite{hayes2009}. The goal of the research program is to devise a formal system that allows linguists to explain the systematic variation in the surface forms (SFs) of a language. 
For instance, consider how the past tense
is expressed in English---a classic case of allomorphy. Comparing \textit{talked} ([\textipa{t\super{h}O:kt}]), \textit{saved} ([\textipa{seIvd}]), and
\textit{acted} ([\textquotesingle\textipa{\ae{}k.tId}]), at least three
pronunciations ([t], [d], [\textipa{I}d]) and at least two spellings (-\textit{ed} and -\textit{d}) can be counted for the past tense morpheme. Traditionally, linguists assume that each morpheme has a single underlying form (UF), a string of phonemes shared across all its contexts \citep{jakobson1948russian}: for the past tense, /-d/. The surface variation is explained by generative phonology via a grammar, a set of rewrite rules \cite{Chomsky1968} or constraint rankings \citep{prince2008optimality} that map the UFs of a sequence of morphemes to the observed SFs.

\begin{figure}[t]
    \centering
\includegraphics[width=\columnwidth]{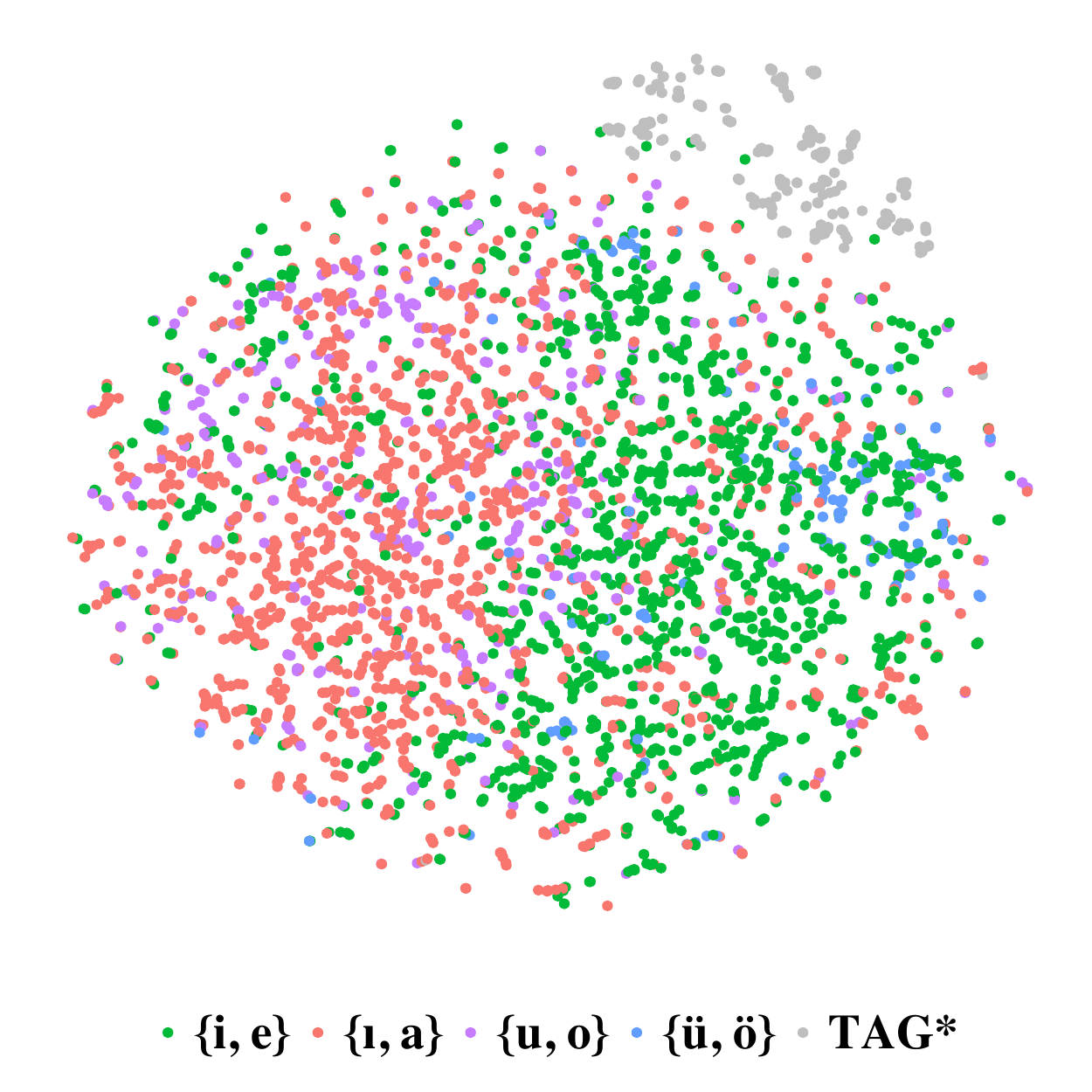}
\caption{Learned underlying representations for Turkish, dimensionality-reduced to $\mathbb{R}^2$ with $t$-SNE. The gray dots show
the suffixes and the colored dots show the stems. 2-way vowel harmony is encoded: back vowels (red and purple) concentrate on the left and front vowels (green and blue) on the right. 
}
\label{fig:urs}
\end{figure}

In this work, we
offer the first end-to-end differentiable version of generative phonology, i.e.\ 
a neural network, trainable by backpropagation, than can perform the same
function that a generative phonology does: To derive the set of attested SFs.
In terms of innovation for linguistic theory, the primary motivation
for this paper is to reconsider the mathematical type of
UFs, which are latent variables (i.e.\ never observed).
What if, rather than being string-valued, we consider
a \emph{more abstract} representation, namely, a
vector in $\mathbb{R}^d$. This would immediately
yield certain advantages: In an end-to-end differentiable
phonology, we can backpropagate through the underlying representations themselves, rather than have a human or machine perform a difficult combinatorial search problem to find the best
string-valued representation. 
We refer to this framework automating the process of being a phonologist as \defn{differentiable phonology}.

In the technical section of the paper, we discuss a series of possible instantiations of differentiable phonology.\footnote{Note that there are many possible neuralizations of generative phonology, several of which are beyond the scope of this paper.} 
Drawing on recent
work in the morphological inflection literature \cite{K17-2001}, we develop neural sequence-to-sequence models to spell out the attested SFs given the word morphemes. 
In particular, we compare three variants relying on different assumptions: a) the presence (or absence) of an intermediate latent variable to bridge between morphemes and SFs, namely UFs with abstract real-valued vector representations; and b) the conditional independence (or dependence) of UFs with respect to SFs.

Moreover, we take time to discuss the impact of these ideas on phonological theory: Despite not being fully interpretable, as traditional underlying strings are, real-valued vector representations can be flexibly compared through cosine similarity. We exhibit learned underlying representations from our differentiable phonology in \cref{fig:urs}. For instance, we find that some classic phonological patterns, e.g. vowel harmony, are immediately evident from this visualization, where morphemes with back and front vowels cluster separately.

Empirically, we provide results on the phonological dataset taken from CELEX2 \cite{baayen1995celex2} and the orthography-based dataset of UniMorph \cite{KIROV18.789}. The task is to predict missing forms in a paradigm (some slots are held out), using the inferred underlying representations. We show that our model of differentiable generative phonology has comparable performance to the latent-string model of \newcite{TACL480} on the tiny CELEX2 dataset. However,
the model scales to the larger UniMorph datasets that are two orders of magnitude larger. Moreover, we find that the best-performing neuralization of differentiable phonology justifies two propositions of generative phonology, viz.\ the presence of UFs and their conditional independence from SFs.

\section{Generative Phonology}\label{sec:generative-phonology}
In this section, we will briefly formalize generative phonology in our own
notation. First, we define the following six sets:
\begin{itemize}
\itemsep0em 
    \item $\Delta = \{d_1, \ldots, d_{|\Delta|}\}$ is the discrete
      alphabet of underlying phonemes.
  \item $\Sigma = \{s_1, \ldots, s_{|\Sigma|}\}$ is the discrete
    alphabet of surface phones.
  \item $M = \{\mu_1, \ldots, \mu_{|M|}\}$ the discrete set of
    abstract morphemes.
  \item $U \subset \Delta^*$ is the set of underlying forms. The
    optimality-theoretic notion of \defn{richness of the base}
    \cite{prince2008optimality} suggests that we should always consider all potential
    forms.
\item $S \subset \Sigma^*$ is the set of realizable surface
  forms.\footnote{Note that many---if not most---of these forms will
    be phonotactically invalid under the language's grammar.}
\item $\tilde{S} \subset S$ is the \emph{observed} set of surface forms that the phonologist has access to to perform their analysis. 
\end{itemize}
Furthermore, let $ s_1\cdots s_{|\surface|} = \surface \in S$ be a \defn{surface form}.
In phonological theory, a surface form is an observed sequence of phonological symbols, e.g., symbols of the International Phonetic Alphabet (IPA). This stands in contrast to the notion of an \defn{orthographic form}, which is the sequence of orthographic symbols used in writing. To see the difference, contrast
the orthographic form \word{talked} and the surface form [\textipa{t\super{h}O:kt}] of the same word.

A surface form $\surface$ may be decomposed, semantically, into a
(variable-length) vector of \defn{abstract morphemes}:\footnote{This term is non-standard in the phonological literature, but we find it useful for our exposition.} 
\begin{equation}
\decomp(\surface) =
\vmu_{\surface} = [\mu_1, \ldots, \mu_k]
\end{equation}
With the term abstract morpheme, we are referring to an abstract unit that carries semantics, but does not have phonological or phonetic content associated with it. 
For each abstract morpheme
$\mu \in M$ in a word, let $\mm_\mu \in U$ be its \defn{underlying
representation}. The concept of the underlying representation is 
commonplace in generative phonology. Using
the terminology of machine learning, an underlying form is a discrete latent variable that helps explain the relation between the surface forms in the lexicon for the same abstract morpheme. 

As a concrete example of our decomposition into abstract morphemes, consider that the English
\word{ran} may be decomposed into its constituent abstract morphemes
\mtag{run} and \mtag{past}. 
Then, the underlying forms for the morphemes are $\mm_{\text{\mtag{run}}} = \text{/\textipa{\*r2}n/}$ and
$\mm_{\text{\mtag{past}}}  = \text{/d/}$.
As a short hand, we will write $\mathbf{m}_{\surface} = [\mm_{\mu_1}, \ldots,
  \mm_{\mu_k}]$.  The underlying forms of the morphemes are stitched
  together to create an underlying form for the entire word. 
  We will write $\uu_{\surface}$ as the underlying form
for the entire surface form. In general, $\uu_{\surface}$ will also be a
latent variable. For instance, the underlying form for [\textipa{\*r\ae{}n}] (\word{ran})
may be /\textipa{\*r2}n\#d/ with a special symbol \# inserted to mark the morphological
boundary. 

A \defn{generative phonology} $\calP : U \rightarrow S$ is a mapping from the
space of underlying forms $U$ to the space of surface forms $S$.
The function $\calP$ may take the form of string rewrite rules,
as suggested by \cite{Chomsky1968} or ordered, violable  constraints, as
suggested by \cite{prince2008optimality}.
Indeed, there many different formalisms
for $\calP$---we will avoid the details of any one specific formalism, taking a broader perspective.
The job of
a generative phonologist is, then, to discover an underlying form
of the words $\under_{\surface}$ for every surface form $\surface$ as well
as to specify the phonology $\calP$ within
their chosen formalism.

What generative phonology is to be preferred? First, it must explain the data: given
the empirical set of surface forms $\tilde{S} \subset S \subset \Sigma^*$ that the
phonologist has access to, the function $\calP$ is chosen,
along with corresponding underlying forms $\under_{\surface}$, such
that $\surface = \calP(\under_{\surface}), \forall \surface \in \tilde{S}$.
Second, the quality of the phonology is evaluated through Occam's razor---simpler phonologies are better than more complex ones, to the extent that both explain the data.
This idea
goes back to \citet[p. 118]{chomsky1955logical}, but has recently been operationalized
with information theory \cite{rasin2016evaluation}. Also, the phonology should
avoid language-specific constraints. In optimality theory, for example,
the set of constraints in the grammar is said to be universal
and languages differ only in their ordering of the constraints. 

Another desideratum is that the phonology $\calP$ can account for the gradient nature of well-formedness judgments of speakers \cite{hayes2008maximum}. This justifies the probabilistic treatments of generative phonology that have been given 
in the literature \cite{jarosz2006richness,apoussidou2006line,eisenstat2009learning,jarosz2013learning,TACL480}.
In broad strokes, these models have treated $\mathbf{u}_{\surface}$ as a string-valued latent
variable. In the fully probabilistic case of \newcite{TACL480}, the goal was to define a joint probability
distribution over surface forms and underlying forms and then to marginalize
out the underlying form as it was never observed.
Working within the framework of string-valued graphical modeling \cite{D09-1011}, one can approximately perform inference despite the $\Delta^*$-valued latent variable using a variant of belief propagation. In contrast, we focus on
a probabilistic generative phonology that has real-valued latent variables as underlying representations. 

\section{The Move to Gradience}
The methodological jump we make in this paper, in contrast to the
discrete latent-variable approach discussed in \newcite{TACL480}, is to treat
the underlying forms as \emph{gradient}, i.e. to treat the underlying forms as a continuous latent variable. This is loosely inspired by the
ideas of gradient symbolic computation \cite[GSC;][]{smolensky2006harmonic}. However,
we emphasize that this connection is loose as GSC creates gradient structures
as a blend of discrete ones, whereas we remove the notion of discreteness
altogether. Much like \newcite{TACL480}, the final form of our model will be a probabilistic latent-variable model. However, in contrast, using the notation in \cref{sec:generative-phonology}, we will define $U = \mathbb{R}^d$.
Thus, each underlying form is a vector in $U$, e.g., 
$\mm_\mtag{run} \in \mathbb{R}^d$ and $\mm_\mtag{past} \in \mathbb{R}^d$. 

\paragraph{Linguistic Justification}
From a theoretical perspective, a more abstract representation of UFs is far from being far-fetched. In the generative framework,
\newcite{kiparsky1968abstract} initiated
the discussion about how much phonological theory should
permit phonological representations to deviate from phonetic reality \citep[ch.\ 6]{kenstowicz2014generative}: conceptualizing contrasts in the underlying forms that never emerge in SFs (so-called `absolute neutralizations').\footnote{Contrary to contextual neutralizations, these are irreversible (they cannot re-emerge in SFs after language change), unstable (they lead to reanalysis of the words involved in a lexical fashion) and non-productive.} The stance that some segments may not be realized phonetically was similarly held by \citet{hyman1970concrete}.

Outside the generative framework, additional arguments have been put forth challenging the idea of discrete underlying forms, and arguing instead in favor of a continuous characterization of phonology. First, this succeeds in grounding phonology on the acoustic and articulatory space of phonetics --a problem known as `naturalness' \citep{jakobson1952preliminaries}. For instance, it explains the parallel between phonetic consonant-vowel co-articulation and phonological assimilation \cite{ohala1990phonetics,flemming2001scalar}. Second, it explains lexical diffusion. Sound changes spread throughout the lexicon incrementally, rather than abruptly as one would expect with discrete phonemes \citep{bybee2007frequency}. Moreover, the rate of change is known to correlate with word frequencies. In order for usage to affect lexical diffusion, the underlying representations must be amenable to vary along a continuum.


In this work, we commit to the assumptions of traditional generative phonology, which exclude considerations of usage and performance from the assessment of competence in a language. However, we abandon the assumption that phonology is discrete in nature. This is not to deny the presence of classical phonemic units at some level of abstraction; rather, we maintain that they can be ``\textit{embedded in a continuous description},'' borrowing the expression of \citet[p. 20]{pierrehumbert2000conceptual}. This leaves open the possibility that phonological constraints and finer-grained phonetic details may co-exist in the same representation \citep{flemming2001scalar}.


%

\begin{figure*}
\centering
\begin{tikzpicture}[thick,scale=0.9, every node/.style={transform shape}]


  \node[latent]                   (m1)      {\Large {\mybox{$\mathbf{m}_\word{talk}$}}} ; %
  \node[latent, right=of m1]       (m2)     {\Large {\mybox{$\mathbf{m}_\word{ed}$}}} ; %
  \node[latent, right=of m2]       (m3)     {\Large {\mybox{$\mathbf{m}_\word{s}$}}} ; %
  \node[latent, right=of m3]       (m4)     {\Large {\mybox{$\mathbf{m}_\word{pad}$}}} ; %

  \node[latent, below=of m1]    (u1)      {\Large {\mybox{$\mathbf{u}_\word{talks}$}}} ; %
   \node[latent, below=of m2]    (u2)      {\Large {\mybox{$\mathbf{u}_\word{talked}$}}} ; %
   \node[latent, below=of m3]    (u3)      {\Large{\mybox{$\mathbf{u}_\word{paded}$}}} ; %
     \node[latent, below=of m4]    (u4)      {\Large {\mybox{$\mathbf{u}_\word{pads}$}}} ; %
     
  \node[obs, below=of u1]    (s1)  {\Large {\myboxgrey{$\surface_\textit{talks}$}}}; %
\node[obs, below=of u2]    (s2)  {\Large {\myboxgrey{$\surface_\textit{talked}$}}}; %
    \node[obs, below=of u3]    (s3)  {\Large {\myboxgrey{$\surface_\textit{padded}$}}}; %
   \node[obs, below=of u4]    (s4)  {\Large{\myboxgrey{$\surface_\textit{pads}$}}}; %
\node[draw=none,rectangle, left=of m1]  (t1) {\text{Morpheme UFs} $\in U = \mathbb{R}^d$} ;
\node[draw=none,rectangle, left=of u1]  (t2) {\text{Word UFs} $\in U = \mathbb{R}^d$} ;
\node[draw=none,rectangle, left=of s1]  (t3) {\text{Word SFs} $\in S = \Sigma^*$} ;



  \edge {m1} {u1} ; %
  \edge {m3} {u1} ; %
  \edge {m1} {u2} ; %
  \edge {m2} {u2} ; %
  \edge {m2} {u3} ; %
  \edge {m4} {u4} ; %
  \edge {m3} {u4} ; %
  \edge {m4} {u3} ; %

  \edge{u1} {s1} ;
  \edge{u2} {s2} ;
  \edge{u3} {s3} ;
  \edge{u4} {s4} ;




\end{tikzpicture}
\caption{Our factorized model can be described as a graphical model that explains the entire lexicon. There are two sets of real-valued latent variables: those for the underlying forms of the morphemes and those for the underlying forms of the words themselves. The picture above shows a fragment of the model that illustrates the relationship among the four forms: \word{talks}, \word{talked}, \word{pads}, \word{padded}.}
\label{fig:factorized}
\end{figure*}
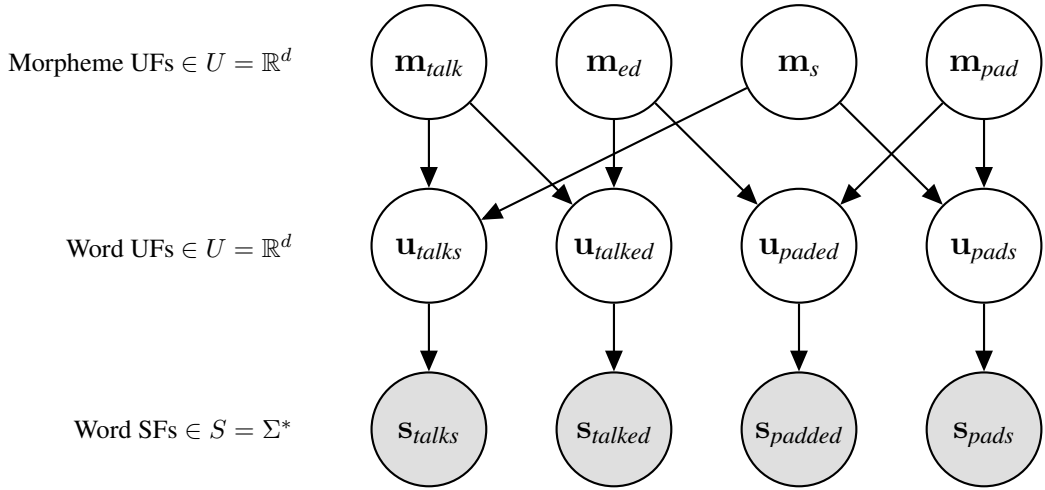
\paragraph{Doing Linguistics Through Backpropagation.}
As mentioned in \cref{sec:generative-phonology}, the goal of the generative phonologist is to posit
underlying forms that help explain the relationships among
the observed surface forms $\tilde{S}$. Even when this problem is posed
probabilistically using latent-variable models, one still has to
contend with a tricky discrete inference problem---a sum over $\Delta^*$.
In contrast, we are going to learn the underlying representations through
backpropagation \cite{rumelhart1985learning}, as they live in a continuous space. The beauty
of this approach is that we can compute the gradient exactly in polynomial time.\footnote{If we introduce continuous latent variables, we will have to rely on approximation.} This means that we can automate the work of a phonologist
using a neural network. 

However, as mentioned above, we will have to live
with the idea that our learned representations are considerably
more abstract than those in traditional phonology. As a consequence, our underlying forms have no direct phonetic interpretation---they are merely vectors in $\mathbb{R}^d$---and thus do not lead to any indirect association with similarly sounding words. However, despite this, our underlying forms can be discussed in relation to their mutual similarity, based on any distance metric (e.g.\ cosine distance). In addition, their most salient properties can be visualized via 
dimensionality reduction (e.g.\ Principal Component Analysis).

\section{A Probabilistic Model for Differentiable Generative Phonology}
We will formulate our differentiable generative phonology as a probabilistic model of the set of observed surface forms $\tilde{S}$ conditioned on the morphological lexicon $M$. We
factorize this distribution as follows:
\begin{align}
  p(\tilde{S} \mid M) &= \prod_{\surface \in \tilde{S}} \pphon(\surface \mid  \mathbf{m}_{\surface}) \\
    &= \prod_{\surface \in \tilde{S}} \prod_{i=1}^{|\surface|} \pphon(s_i \mid \surface_{< i}, \mathbf{m}_{\surface}) 
\end{align}
where, again, we define $\mathbf{m}_{\surface} = [\mm_1; \ldots; \mm_{|\vmu|}] \in \mathbb{R}^{|\vmu| \times d}$.
To complete the generative model,\footnote{The model is generative only in so as we generate all the sounds. We condition on the structure of the lexicon itself.} we further
define a Gaussian prior over the morphemic underlying forms
\begin{equation}
    p(\mm_i \mid \mu_i) = {\cal N}(\mathbf{0}, I).
\end{equation}
As described in \cref{sec:generative-phonology}, generative phonology also presupposes a factorization between the surface form and the underlying form. We encode this factorization as follows
\begin{align}\label{eq:factorization}
    \pphon(s_i, &\mid \surface_{< i}, \mathbf{m}_{\surface}) = \\
    &\int \psr(s_i, \mid \surface_{< i}, \uu_{\surface} )\, \pur(\uu_{\surface} \mid \surface_{<i}, \mathbf{m}_{\surface}) \, \mathrm{d}\uu_{\surface} \nonumber
\end{align}
where we have added a latent, real-valued vector $\uu_{\surface} \in \mathbb{R}^d$ that corresponds
to the underlying form in phonology. This completes the move
from a discrete, string-valued underlying form, to a gradient, real-valued one. Note
that the subscripts on $\psr$ and $\pur$ stand for surface representation
and underlying representation, respectively.

This general model allows for several different parameterizations. 
Here, we will discuss three different variants: two make use of the
factorization in \cref{eq:factorization}, i.e. keeping the spirit 
of an underlying form, but make different assumptions regarding the conditional independence of $\uu_{\surface}$ from $\surface$, as described in \cref{ssec:factmod}. The third variant, outlined in \cref{ssec:nour} directly parameterizes the joint distribution $\pphon$, removing the underlying form as an explicit latent variable. To perform inference over the latent-variable models, we resort to the variational approximation in \cref{ssec:varinf}.


\subsection{Factorized Model}
\label{ssec:factmod}
Our factorized model, which assumes an intermediate latent variable for underlying forms, is depicted as a directed graph in \cref{fig:factorized}.
\paragraph{Defining $\psr$.}
We first define the part of the neural network that spells out the
surface form conditioned on the underlying representation $\uu_\surface$.
We propose that this is a conditional long short-term memory \cite[LSTM;][]{hochreiter1997long} decoder,
similar to the one used in \newcite{NIPS2014_5346} for neural machine
translation. This is a recurrent neural network that conditions the decoder on a fixed-length vector. More formally, we define this network
as the following
\begin{equation} \label{eq:psr}
  \psr(s_i \mid \surface_{< i}, u_{\surface}) = \textrm{softmax}\left(f([\mathbf{h}^\textrm{(dec)}_i; \uu_{\surface}]) \right)
\end{equation}
where $\mathbf{h}^\textrm{(dec)}_i \in \mathbb{R}^d$ is the hidden state of a decoder LSTM reading the embedding of the previous character and $f$ is a two-layer neural network
\begin{equation} \label{eq:ffn}
    f(\mathbf{h}^\prime_i) = {V}\,\textrm{tanh}({W}\,\mathbf{h}^\prime_i)
\end{equation}
where $W \in \mathbb{R}^{2d \times 2d}$ and $V \in \mathbb{R}^{|\Sigma| \times 2d}$.
By preventing the LSTM access of $\mathbf{m}_{\surface}$, $\uu_{\surface}$ is forced to encode enough information---together with a hidden state  $\mathbf{h}^\textrm{(dec)}_i$ encoding past characters---to output the correct character through a feed-forward network $f$.
The exact construction of the underlying form $\uu_{\surface}$ 
will depend on the distribution $\pur$, which in turn will hinge upon
the nature of the attention mechanism we use to compose the morpheme representations. 

\paragraph{Position-Independent UF.}
Traditionally, in generative phonology, the underlying representation
is taken to be independent of how the word is spelled-out: within our model, we express this
as the following conditional independence assumption:
\begin{align} \label{eq:posindur}
\pur(\uu_{\surface} \mid \surface_{< i}, \mathbf{m}_{\surface}) = \pur(\uu_{\surface} \mid \mathbf{m}_{\surface}) 
\end{align}
In this case, we may derive a version on the model where the integral
is moved outside the second product:
\begin{align}
&\prod_{\surface \in S} \prod_{i=1}^{|\surface|} \int \psr(s_i \mid \surface_{< i}, \uu_{\surface} )\, \pur(\uu_{\surface} \mid \mathbf{m}_{\surface}) \,\textrm{d}\uu_{\surface} \\
&= \prod_{\surface \in S} \int \left( \prod_{i=1}^{|\surface|} \psr(s_i \mid \surface_{< i}, \uu_{\surface} )\right)  \pur(\uu_{\surface} \mid \mathbf{m}_{\surface}) \,\textrm{d}\uu_{\surface} \nonumber
\end{align}
Using this simplified model, we define the position-independent UF distribution as the following
\begin{align}
\pur(\uu_{\surface} \mid \mathbf{m}_{\surface}) = {\cal N}\left(\frac{1}{|\vmu_{\surface}|} \sum_{i=1}^{|\vmu_{\surface}|} \mm_i, I \right).
\end{align}
This is a simple averaging of the underlying forms $\mm_i$. Note
that there are many possible distributions over underlying forms $\uu_{\surface}$.

\paragraph{Position-Dependent UF.}
Contrary to the customary practice in generative phonology, we also
consider a position-dependent underlying representation. This means
that the underlying representation of the word changes as the SF is being spelled out. We define such a position-dependent UF
as
\begin{align}
  \pur(\uu_{\surface} \mid \surface_{< i}, \vmu_{\surface}) = {\cal N}\left(\sum_{i=1}^{|\vmu_{\surface}|} \alpha_i \mm_i,  I \right).
\end{align}
The attention weights are defined as
\begin{equation} \label{eq:attn}
\alpha_i = \dfrac{\exp({\vh^{\textrm{(dec)}}_i}^{\top}\, {T}\, \mm_i)}{\sum_{j=1}^{|\vmu_{\surface}|} \exp({\vh^{\textrm{(dec)}}_j}\, {T}\, \mm_{j})}
\end{equation}
where ${T} \in \mathbb{R}^{d \times d}$ is a learned parameter matrix.
This model is equivalent to a Bayesian version of the 
classic soft attention model of \newcite{D15-1166}. Why position-dependent underlying representations?
Based on the machine translation literature, it could potentially yield better performance since it is a more expressive model. However, foreshadowing, as we show in the empirical portion of this paper, it underperforms position independent UF by a large margin.

\subsection{Removing the Underlying Representation}
\label{ssec:nour}
The final variant we consider is a joint model of phonology:
that is, we directly condition the distribution
over surface forms on the morpheme representations.\footnote{This variant may be interpreted as a way to implement \citet{hooper1976introduction}'s `true generalization condition,'  according to which speakers formulate phonological rules that operate directly at the surface level.} The implementation we choose takes inspiration
from hard attention
\begin{align}
  \pphon(s_i \mid \surface_{< i}, \mathbf{m}_{\surface}&) =  \\
  &\sum_{i=1}^{|\vmu_{\surface}|} \alpha_i\, \textrm{softmax}\left(f([\vh_i; \mm_i])\right) \nonumber
\end{align}
where $\alpha_i$ is defined as above.
Contrary to the models in \cref{ssec:factmod}, here we lose the notion of a distinct underlying form. 

\subsection{Neural Variational Inference}
\label{ssec:varinf}
In the case of the factorized models, we treat
the underlying representation $\us$ of a given surface
form $\surface$ as as latent variable that we marginalize out.
This introduces some computational complications. Namely, the integral of \cref{eq:factorization} is intractable to compute. For this reason,
we turn to an approximation: variational inference. 
In particular, we optimize the evidence lower bound (ELBO) following a derivation similar to \newcite{kingma2013auto}: 

\begin{align} \label{eq:elbo}
&\log p(\tilde{S} \mid M) \\ \nonumber
&= \log\prod_{\surface \in \tilde{S}} \prod_{i=1}^{|\surface|} \int \psr(s_i, \mid \uu_{\surface} )\, \pur(\uu_{\surface} \mid \mathbf{m}_{\surface}) \textrm{d}\uu_{\surface}\\ \nonumber
&= \sum_{\surface \in \tilde{S}} \sum_{i=1}^{|\surface|} \log \int \psr(s_i, \mid \uu_{\surface} )\, \pur(\uu_{\surface} \mid \mathbf{m}_{\surface}) \textrm{d}\uu_{\surface}\\ \nonumber
&\geq \sum_{\surface \in \tilde{S}} \sum_{i=1}^{|\surface|} \int \log \psr(s_i, \mid \uu_{\surface} )\, \pur(\uu_{\surface} \mid \mathbf{m}_{\surface}) \textrm{d}\uu_{\surface}\\
\end{align}
Note that $\surface_{< i}$ is omitted in $\psr$ and $\pur$ for simplicity.
The integral is estimated by a single sample from $\pur$ with the re-parameteriation trick \cite{kingma2013auto}, which makes the sample differentiable. In practice, for position-dependent UF, a sample has the following form
\begin{align}
\hat{\uu}_{\surface} = \sum_{i=1}^{|\vmu_{\surface}|} \alpha_i \mm_i +\veps
\end{align}
where $\veps \sim {\cal N}\left(\mathbf{0},  I \right)$. For position-independent UF, the one sample is drawn as $\hat{\uu}_{\surface} = \frac{1}{|\vmu_{\surface}|} \sum_{i=1}^{|\vmu_{\surface}|} \mm_i + \veps$.

\begin{figure*}
\centering
\begin{adjustbox}{width=1.7\columnwidth}
\includegraphics{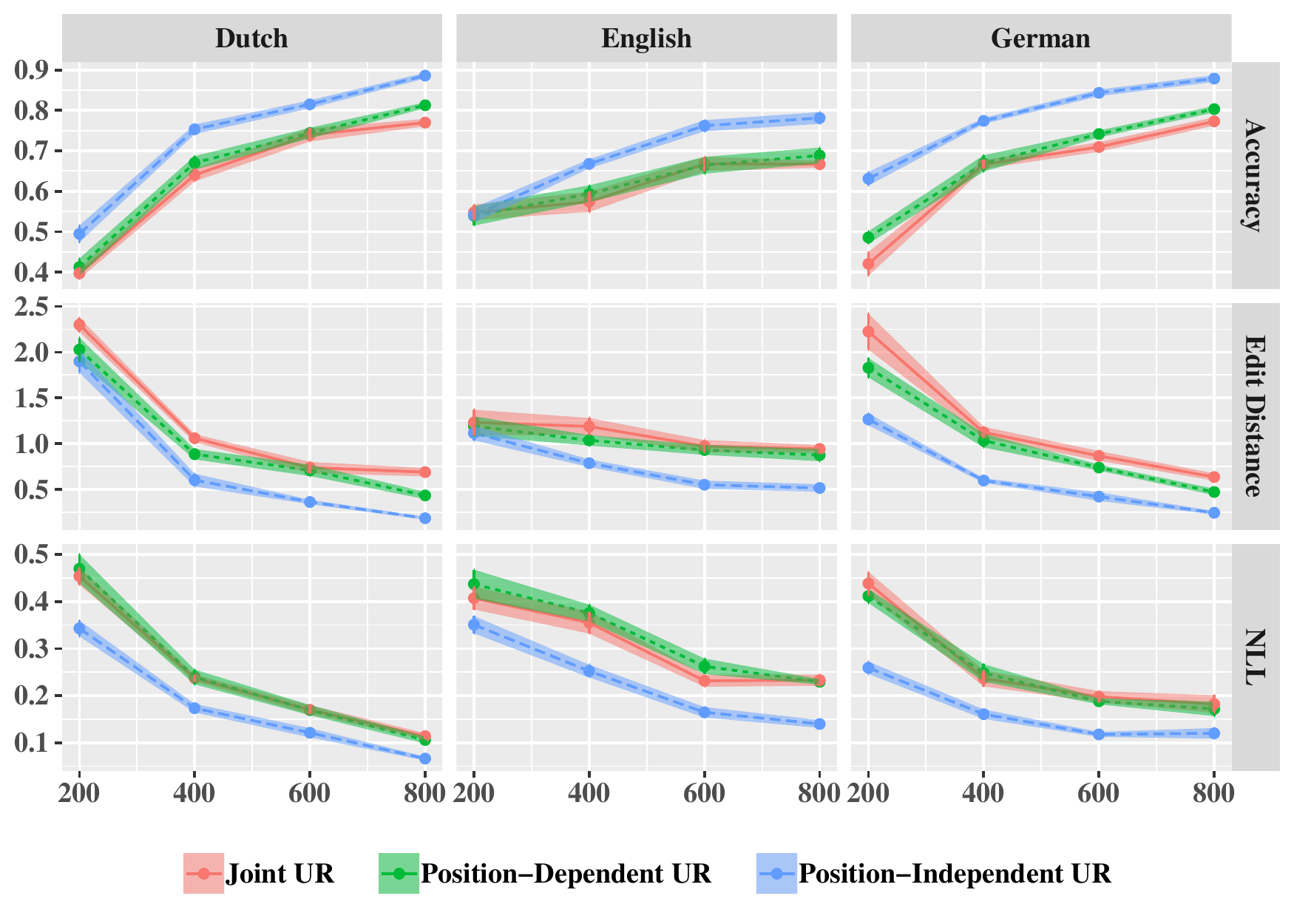}
\end{adjustbox}
\caption{Accuracy ($\uparrow$), edit distance ($\downarrow$), and surprisal (\textsc{nll} $\downarrow$) on 3 languages of CELEX2. The y-axis represents the metric value, and the x-axis the amount of training examples. Shaded areas indicate one standard deviation for each estimate. Again, the Position-Independent UF outperforms both competitors by significant margins.}
\label{fig:celex}
\end{figure*}

\section{Experiments}
In the experimental part of this work, we assess to which extent our differentiable generative phonology can predict held-out surface forms (and orthographic forms) in word paradigms given a sequence of abstract morphemes. Note that our evaluation setup differs from the established practice in linguistics, where a system of rules or constraint rankings---combined with underlying forms---is favored over the alternatives if it better explains the full set of data and possibly satisfies other desiderata, such as simplicity (see \cref{sec:generative-phonology}). Instead, our setup based on blind testing quantifies the ability of any model of generative phonology to \textit{generalize} to held-out forms.\footnote{Measuring generalization averts theory-internal considerations but implicitly selects for both accurate and simple models, as it indicates lack of over-fitting to the data.} In particular, we rely on three disjoint sets of data points: a training set to perform inference, a development set for fine-tuning hyper-parameters, and an evaluation set to test the model predictions.

\begin{table*}[t]
\centering
\begin{tabular}{>{\scshape}l rrr rrr rrr} \toprule
& \multicolumn{3}{c}{Position-Independent UF} & \multicolumn{3}{c}{Position-Dependent  UF} & \multicolumn{3}{c}{Joint Model UF} \\
\cmidrule(lr){2-4} \cmidrule(lr){5-7} \cmidrule(lr){8-10}
& \textbf{ACC $\uparrow$} & \textbf{MLD $\downarrow$} & \textbf{NLL $\downarrow$} & \textbf{ACC $\uparrow$} & \textbf{MLD $\downarrow$} & \textbf{NLL $\downarrow$} & \textbf{ACC $\uparrow$} & \textbf{MLD $\downarrow$} & \textbf{NLL $\downarrow$} \\ \midrule
 ara & 98.7 & 0.039 & 0.005 & 79.6 & 0.823 & 0.059 & 79.1 & 0.812 & 0.056 \\
 bul & 96.8 & 0.060 & 0.016 & 84.3 & 0.389 & 0.048 & 81.0 & 0.555 & 0.054 \\
 ces & 94.3 & 0.099 & 0.024 & 84.6 & 0.437 & 0.057 & 82.3 & 0.660 & 0.063 \\
 cym & 94.6 & 0.108 & 0.032 & 89.3 & 0.257 & 0.050 & 79.1 & 0.594 & 0.070 \\
 dan & 65.6 & 0.402 & 0.064 & 55.9 & 1.107 & 0.113 & 57.3 & 1.249 & 0.104 \\
 deu & 96.4 & 0.076 & 0.016 & 67.6 & 1.383 & 0.124 & 67.5 & 1.441 & 0.114 \\
 eng & 90.9 & 0.179 & 0.045 & 44.8 & 1.819 & 0.212 & 37.7 & 2.370 & 0.238 \\
 est & 92.5 & 0.249 & 0.021 & 75.2 & 0.681 & 0.062 & 82.5 & 0.609 & 0.045 \\
 eus & 72.8 & 0.953 & 0.127 & 54.2 & 1.646 & 0.179 & 57.2 & 1.476 & 0.162 \\
 fas & 99.7 & 0.004 & 0.001 & 92.4 & 0.135 & 0.014 & 93.0 & 0.119 & 0.012 \\
 fra & 98.6 & 0.030 & 0.006 & 86.9 & 0.266 & 0.038 & 86.1 & 0.339 & 0.039 \\
 gle & 85.6 & 0.510 & 0.043 & 54.8 & 1.410 & 0.121 & 29.2 & 4.550 & 0.246 \\
 heb & 94.5 & 0.070 & 0.032 & 63.2 & 0.598 & 0.181 & 56.7 & 0.814 & 0.212 \\
 hin & 100.0 & 0.001 & 0.000 & 94.8 & 0.074 & 0.007 & 93.0 & 0.104 & 0.010 \\
 hye & 99.6 & 0.007 & 0.001 & 93.6 & 0.180 & 0.019 & 91.8 & 0.263 & 0.023 \\
 ita & 99.4 & 0.014 & 0.003 & 77.0 & 0.530 & 0.065 & 76.5 & 0.561 & 0.063 \\
 lav & 97.9 & 0.038 & 0.010 & 55.0 & 1.234 & 0.105 & 52.5 & 1.419 & 0.116 \\
 nld & 96.8 & 0.069 & 0.017 & 81.8 & 0.573 & 0.072 & 80.2 & 0.727 & 0.069 \\
 pol & 92.6 & 0.172 & 0.031 & 70.5 & 0.916 & 0.115 & 74.2 & 0.948 & 0.096 \\
 por & 99.7 & 0.006 & 0.002 & 94.3 & 0.109 & 0.019 & 93.1 & 0.158 & 0.020 \\
 ron & 77.8 & 0.806 & 0.065 & 61.7 & 1.422 & 0.128 & 53.3 & 1.948 & 0.159 \\
 rus & 93.7 & 0.198	& 0.025 & 60.9 & 1.909 & 0.160 & 66.3 & 1.741 & 0.122 \\
 spa & 99.5 & 0.014 & 0.002 & 86.1 & 0.346 & 0.047 & 84.9 & 0.388 & 0.040 \\
 sqi & 98.7 & 0.028 & 0.006 & 86.7 & 0.362 & 0.031 & 89.5 & 0.272 & 0.026 \\
 swe & 92.7 & 0.142 & 0.034 & 72.4 & 0.992 & 0.119 & 67.2 & 1.604 & 0.135 \\
 tur & 89.8 & 0.301 & 0.013 & 83.7 & 0.550 & 0.026 & 82.7 & 0.653 & 0.027 \\
 ukr & 85.7 & 0.235 & 0.076 & 69.8 & 0.786 & 0.128 & 78.1 & 0.505 & 0.105 \\
 urd & 99.4 & 0.012 & 0.003 & 97.8 & 0.079 & 0.009 & 91.5 & 0.295 & 0.027 \\ \midrule
\textit{average} & \textbf{93.0}	& \textbf{0.172}	& \textbf{0.026} & 75.7 & 0.750 & 0.082 & 73.7 & 0.971 & 0.088 \\
\bottomrule
\end{tabular}
\caption{Accuracy (\textsc{acc} $\uparrow$), edit distance (\textsc{mld} $\downarrow$), and surprisal (\textsc{nll} $\downarrow$) on 28 languages of UniMorph 2.0. The Position-Independent UF outperforms both competitors, often by very large margin.}
\label{tab:unimorph}
\end{table*}

\subsection{Datasets}
We experiment on two datasets, CELEX2 \citep{baayen1995celex2} for phonetic surface forms and UniMorph 2.0 \citep{KIROV18.789} for orthographic forms. The choice of the second dataset highlights the computational advantages of gradient underlying forms. In contrast to models with string-valued latent variables, our model can scale to morphological lexica that contain a number of forms in the same order of magnitude as natural languages. Indeed, \newcite{TACL480}'s largest lexicon covered only 1000 forms. Thus, this work provides the first large-scale induction of underlying forms with a computational model. In what follows, we provide additional details on the datasets. 

\textbf{CELEX 2} \citep{baayen1995celex2} provides surface forms and token counts for 3 languages: Dutch, English, and German. In particular, we use a subcorpus created by \newcite{TACL480}, which contains 1000 nouns and verbs per language, and focuses on voicing patterns (such as final obstruent devoicing and voicing assimilation). The phonological annotation is limited to segmental features, and thus ignores suprasegmental phenomena such as prosody. As a training set, again following \newcite{TACL480}, we consider several subsets of the subcorpus, sampling without replacement $k$ instances based on the distribution defined by normalized token counts. In particular, we consider $k \in \{200, 400, 600, 800\}$. In this dataset, each word has at most 2 abstract morphemes, consisting in a stem and a (possibly empty) affix.

\textbf{UniMorph 2.0} \cite{KIROV18.789} is a dataset of vast proportions and covering a wide spectrum of languages. Moreover, it is the most widespread benchmark for morphological inflection \citep{vylomova2020sigmorphon}. All data are taken from the UniMorph project.\footnote{\url{https://unimorph.github.io}} Specifically, 
we look at the following 28 languages: Albanian, Arabic, Armenian, Basque, Bulgarian, Czech, Danish, Dutch, English, Estonian, French, German, Hebrew, Hindi, Irish, Italian, Latvian, Persian, Polish, Portuguese, Romanian, Russian, Spanish, Swedish, Turkish, Ukrainian, Urdu and Welsh. The languages come from 4 stocks (Indo-European, Afro-Asiatic, Finno-Ugric and Turkic) with Basque, a language isolate, included as well. They represent a reasonable degree of typological diversity. We lament that the Indo-European family is overrepresented in the UniMorph dataset. However, within the Indo-European family, we consider a diverse set of genera: Albanian, Armenian, Slavic, Germanic, Romanace, Indo-Aryan, Baltic and Celtic. Contrary to CELEX2, UniMorph 2.0 contains only orthographic strings and currently lacks a phonemic transcription. 

\subsection{Hyperparameters}
The dimension of morpheme embeddings $\mm$, underlying form embeddings $\uu$, surface character embeddings $\surface$, and the hidden size of the 1-layer decoder LSTM are $d = 200$. Therefore, the learnable parameters for the feed-forward network in \cref{eq:ffn} are ${W}\in\mathbb{R}^{400\times400}$ and ${V}\in\mathbb{R}^{|\Sigma|\times400}$, and for the attention mechanism in \cref{eq:attn} are ${T}\in\mathbb{R}^{200\times200}$. We additionally apply 0.2 dropout \citep{srivastava2014dropout} to all embeddings. The model is trained with Adam \citep{kingma2014adam} and the learning rate is 1e-3. We halve the learning rate whenever the development loss does not improve and we stop early when learning rate drops below 1e-5. Finally, we wait 10 epochs before dropping the learning rate in the CELEX2 experiments.

\subsection{Evaluation Metrics}
We consider 3 evaluation metrics: (i) surprisal (the
negative log probability normalized by length) of the held-out surface or orthographic forms (NLL), 
(ii) accuracy of the 1-best prediction (ACC) and (iii) edit distance of the 1-best prediction (MLD). Given the unfeasibly many possible splits of CELEX2 with $k$ training examples, for each desired size we estimate the expected performance through 10 random samples. In particular, we report the sample mean and the standard deviation of each metric.




\section{Results and Discussion}
We present the results for surface form prediction on CELEX2 in \cref{fig:celex} and the results for orthographic form prediction on UniMorph 2.0 in \cref{tab:unimorph}. Of the three neural models we compare, we find that the position-independent performs the best. This behavior is consistent across both datasets, 3 evaluation metrics (accuracy, edit distance, and surprisal), and 4 sizes of training sets (for CELEX2). For UniMorph 2.0, all the differences are significant under a paired-permutation test with $p < 0.05$. Indeed, this result is quite strong as it holds for every one of our 28 languages.

The model with position-independent UFs is superior to the second-best model with position-dependent UFs as it increases its accuracy by 17.3 absolute points (+23\%), reduces its edit distance by 0.578 points (-77.07\%), and reduces its surprisal by 0.046 points (-56.10\%). This shows the importance of independence assumptions of UFs with respect to SFs. What is more, the joint model without UFs achieves the worst scores, with an additional gap in performance. This result showcases the need for latent variables bridging between abstract morphemes and SFs, which correspond to gradient UFs in our formulation. While both these findings confirm received wisdom from the literature in generative phonology, the preeminence of the variant with position-independent UFs is surprising from the computational perspective: Attention-based models are generally superior to those without attention in morphological tasks \citep[\textit{inter alia}]{aharoni-goldberg-2017-morphological,wu-etal-2018-hard,wu-cotterell-2019-exact}.

The results suggest that a mode that encodes a single underlying
representation is best for modeling the phonological and orthographic forms. If one accepts the connection between our real-valued latent variable and the traditional underlying forms of generative phonology, these results provide support for the idea that an underlying form is useful for explaining the surface forms in the lexicon.

\section{Conclusion}
We have presented a differentiable and probabilistic version of generative phonology. Instead of having string-valued underlying forms, as in traditional generative phonology, we have relaxed this formalism to one where the underlying forms are real-valued vectors. Using sequence-to-sequence neural models that have become standard in natural language processing (NLP), our differentiable generative phonology spells outs the words from this latent vector space. Since our model can be learned in an end-to-end fashion, underlying forms are discovered automatically rather than being posited by linguists. While lacking a phonetic interpretation, these forms can be compared with respect to their geometric distance, revealing phonological patterns such as vowel harmonization. 

The empirical portion of our paper conducts experiments on 3 languages from the CELEX2 dataset for surface form prediction and 28 languages from the  2.0 dataset for morphological inflection. We find that, under three metrics, our model of differentiable phonology achieves performances comparable with previous work in the small-data regime, but scales to many more forms in the large-scale setting. Finally, we compare several variants of our model, lending credibility to two conjectures of generative phonology: that the variation in SFs is mediated by UFs, and that these are conditionally independent from SFs.
All of our code, models and results may be found at \url{https://github.com/shijie-wu/neural-transducer}.

\bibliography{gradient-phonology}
\bibliographystyle{acl_natbib}

\end{document}